\documentclass[letterpaper, 10 pt, conference]{ieeeconf}

\IEEEoverridecommandlockouts
\usepackage[utf8]{inputenc}
\usepackage{amsmath}

\usepackage{amsfonts}
\usepackage{amsthm}
\usepackage{amssymb}
\usepackage{graphicx}
\usepackage{booktabs}
\usepackage{url}
\usepackage{nicefrac}

\usepackage{xcolor}

\usepackage{algorithm}
\usepackage{algpseudocode}
\usepackage{cite}
\title{\textbf{Collaborative Conversation in Safe Multimodal\\Human-Robot Collaboration}}
\author{Davide Ferrari, Andrea Pupa and Cristian Secchi
\thanks{D. Ferrari, A. Pupa and C. Secchi are
    with the Department of Sciences and Methods of Engineering,
    University of Modena and Reggio Emilia, Italy  {\tt\small
    \{davide.ferrari95, andrea.pupa,
    cristian.secchi\}@unimore.it}}
}

\newcommand{\q}{\mathbf{q}}
\newcommand{\dq}{\dot{\mathbf{q}}}
\newcommand{\uin}{\mathbf{u}}
\newcommand{\R}{\in\mathbb{R}}
\newcommand{\traj}{\q_{des}(t)}
\newcommand{\trajs}{\q_{des}(s(t))}
\newcommand{\dtraj}{\dot{\q}_{des}(t)}
\newcommand{\dtrajs}{\q'_{des}(s(t))}
\newcommand{\dqs}{\q'(s)}

\begin{document}

    \maketitle
    
    \begin{abstract}

        In the context of Human-Robot Collaboration (HRC), it is crucial that the two actors are able to communicate with each other in a natural and efficient manner. The absence of a communication interface is often a cause of undesired slowdowns. On one hand, this is because unforeseen events may occur, leading to errors. On the other hand, due to the close contact between humans and robots, the speed must be reduced significantly to comply with safety standard ISO/TS 15066.
        In this paper, we propose a novel architecture that enables operators and robots to communicate efficiently, emulating human-to-human dialogue, while addressing safety concerns. This approach aims to establish a communication framework that not only facilitates collaboration but also reduces undesired speed reduction. Through the use of a predictive simulator, we can anticipate safety-related limitations, ensuring smoother workflows, minimizing risks, and optimizing efficiency. The overall architecture has been validated with a UR10e and compared with a state of the art technique. The results show a significant improvement in user experience, with a corresponding 23\% reduction in execution times and a 50\% decrease in robot downtime.
    
    \end{abstract}

	\section{Introduction}\label{sec:intro}



    Human-Robot Collaboration (HRC) is rapidly gaining importance in various sectors of modern society, such as industry \cite{Matheson2019robotics} \cite{Vil18Mechatronics}, medicine \cite{Haidegger2022IEEE} \cite{Shen2021IEEE} and elder assistance \cite{Mis20IEEE} \cite{Lima2022IEEE}.
    %
    %
    %
    %
    The success of HRC heavily relies on effective communication, akin to the significance of communication in Human-Human Collaboration (HHC) \cite{Mojtahedi2017Frontiers}.
    Ensuring seamless information sharing, accurate task execution in shared workspaces, and clear expression of intentions or requests between human and robot team members is paramount. In safety-focused scenarios, proactive communication becomes even more crucial, serving
    as one of the mechanisms to anticipate and avoid potential hazards or inefficiencies,
    fostering smooth collaboration while minimizing risks.
    %
    %
    %
    %
    While HRC shows promise in enhancing efficiency and safety across domains, existing approaches often struggle to achieve the dynamic, bidirectional, and proactive communication commonplace in human interactions, particularly in safety-related aspects. Communication in \cite{Grushko2021Sensors} and \cite{Chadalavada2020Robotics} enhances coexistence and trust, providing operators with insights into the intentions of robots. Meanwhile, \cite{Rosen2020IROS} and \cite{LIU2020CIRP} employ a multimodal communication approach, enabling unidirectional commands from the operator to the robot, enhancing safety by reducing misunderstandings.
    %
    %
    %
    %
    However, unidirectional communication lacks the ability for the robot to provide crucial feedback, as seen in \cite{Ferrari2022ICRA}, which introduces bidirectional communication where the robot can request safety constraint relaxation using Control Barrier Functions (CBFs). This approach ensures that the robot can express its needs to the operator, enhancing both safety and task efficiency.
    %
    %
    %
    %
    Despite these efforts, existing approaches often prioritize communication or safety separately, leading to a critical gap where safety is not explicitly integrated into communication strategies. Approaches like \cite{benzi2023energy, ferraguti2020control, palleschi2021fast, lucci2020combining} prioritize safety but lack integration with dedicated communication channels, turning safety into a barrier that may cause the robot to stop or slow down without explicitly guaranteeing compliance with ISO/TS 15066 safety standards \cite{isots}.
    This gap underscores the need for further research and development to seamlessly integrate safety and communication, meeting regulatory requirements and ensuring safe collaboration in diverse applications.
    In this article, we introduce an innovative HRC approach that prioritizes natural conversation while utilizing a predictive simulator to proactively anticipate potential safety issues, all while adhering to ISO safety standards. Our architectural framework views human and robot team members as equals, enabling an intuitive bidirectional communication to facilitate dynamic information exchange, closely resembling human-to-human conversations. Through bidirectional communication, the system engages with the user to find solutions, minimizing risks while maintaining high efficiency and ensures smoother workflows avoiding disruptions and preemptively addressing safety concerns.

    Thus, the contributions of this paper are:
    
    \begin{itemize}
        \item An innovative architecture for Bidirectional Multimodal Communication that enables natural dialogues between humans and robots.
        \item A predictive simulator and the seamless integration of safety considerations into the conversation, proactively addressing potential slowdowns or blockages to enhance collaboration efficiency.
        \item An experimental validation by comparing our proposed architecture to the state-of-the-art, where safety functions as a low-level layer and is not an integral part of communication.
    \end{itemize}

    This paper is structured as follows: in Section II, we address the Problem Statement. Section III presents the Proposed Architecture, while Section IV delves into the Safety Layer and the Predictive Simulator. Experimental Validation, Implementation Details, and Analysis of the Results are discussed in Section V, and in Section VI, we draw our Conclusions.

    \section{Problem Statement}\label{sec:problem_statement}
    Consider a Human-Robot Collaboration scenario, where a 6-degree-of-freedom (6-DOF) velocity-controlled manipulator, modeled as:
    \begin{equation}
        \dq = \uin
    \end{equation}
    where $\dq\R^n$ denotes the joints velocities and $\uin\R^n$ represents the controller input, is required to cooperate and establish communication with a human operator to achieve a common objective.
    In this context, humans and robots collaborate to execute a cooperative assembly task within a shared workspace. The robot's primary role is to assist the operator to complete the task efficiently. Utilizing a set of sensors, real-time monitoring of both the human operator's position and the objects in the workspace becomes possible. This real-time monitoring forms the basis for planning safe trajectories $\traj\R^n$, which originate from an initial configuration $\q_{des}(t_i)=\q_i\R^n$ and extend to a desired final configuration $\q_{des}(t_f)=\q_f\R^n$. These trajectories must adhere to the ISO/TS 15066 regulations \cite{isots}, which imposes constraints on the maximum speed in the direction of the operator \cite{pupa2023chapter}:
    \begin{equation}
    \begin{split}
        v_{rh}(t)\le &\sqrt{v_h(t)^2+(a_{max}T_r)^2-2a_{max}K(t)}+\\
        &-a_{max}T_r -v_h(t),
        \label{eq:vellimit}
        \end{split}
    \end{equation}
    where $K(t) = C + Z_d + Z_r-S_p(t)$.
    $v_{{rh}}(t) \R$ and \mbox{$v_{h}(t) \R$} are the scalar velocity of the robot towards the human operator and the scalar velocity of the human operator towards the robot, respectively. $a_{max} \R$ is the maximum deceleration and $T_r\R$ is the robot reaction time. $C$ is the intrusion distance, i.e. the distance that a part of the body can intrude into the sensing field before it is detected, while $Z_d$ and $Z_r$ are the position uncertainties of the human operator inside the workspace and of the robot system, respectively. Lastly, $S_p$ represents the protective separation distance.

    To ensure compliance with safety standards while maintaining the overall path integrity, a strategy is employed that explicitly isolates the velocity magnitude along the trajectory. This is achieved through a path-velocity decomposition method, involving the manipulation of the derivative $\dot{s}$ of the curvilinear abscissa $s$ that parameterizes the geometric path $\trajs$ as follows:
   \begin{align}
        & &\traj &= \trajs &&t\in \begin{bmatrix}t_i, t_f\end{bmatrix},
        \label{eq:q_s}
        \\
        & &\dtraj &= \dtrajs\dot{s} &&t\in \begin{bmatrix}t_i, t_f\end{bmatrix},    
        \label{eq:q_s_dot}
    \end{align}

    
    However, since all robot tasks must be implemented safely, this often translates into inefficient behavior, causing undesired slowdowns or blockages. In a HHC scenario, the two actors would anticipate potential issues by discussing and exchanging valuable information to avoid them. To emulate this, it is first necessary to implement a multimodal conversation architecture capable of handling multiple communication channels and enabling the exchange of information in a simple, natural, and dynamic manner.
    Secondly, the robot needs to be equipped with an algorithm that allows it to assess and predict the possible emergence of such issues, enabling it to converse with the operator and find a solution.

    This work aims to address these issues by enabling multimodal conversations, which, through effective communication and predictive simulation, aim to prevent the emergence of potential future problems.
    To achieve this, we propose a safe-integrated architecture that:
    \begin{itemize}
        \item enables multimodal conversations between humans and robots to ensure efficient and natural information exchange.
        \item anticipates the emergence of potential issues or slowdowns.
        \item ensures compliance with regulations during the execution of the robot's trajectories.
    \end{itemize}

    
\section{Proposed Architecture}\label{sec:proposed architecture}
    


   \begin{figure}
        \vspace{5pt}
        \centering
        \includegraphics[trim={9cm 4.5cm 8.5cm 2.5cm}, clip, width=\linewidth]{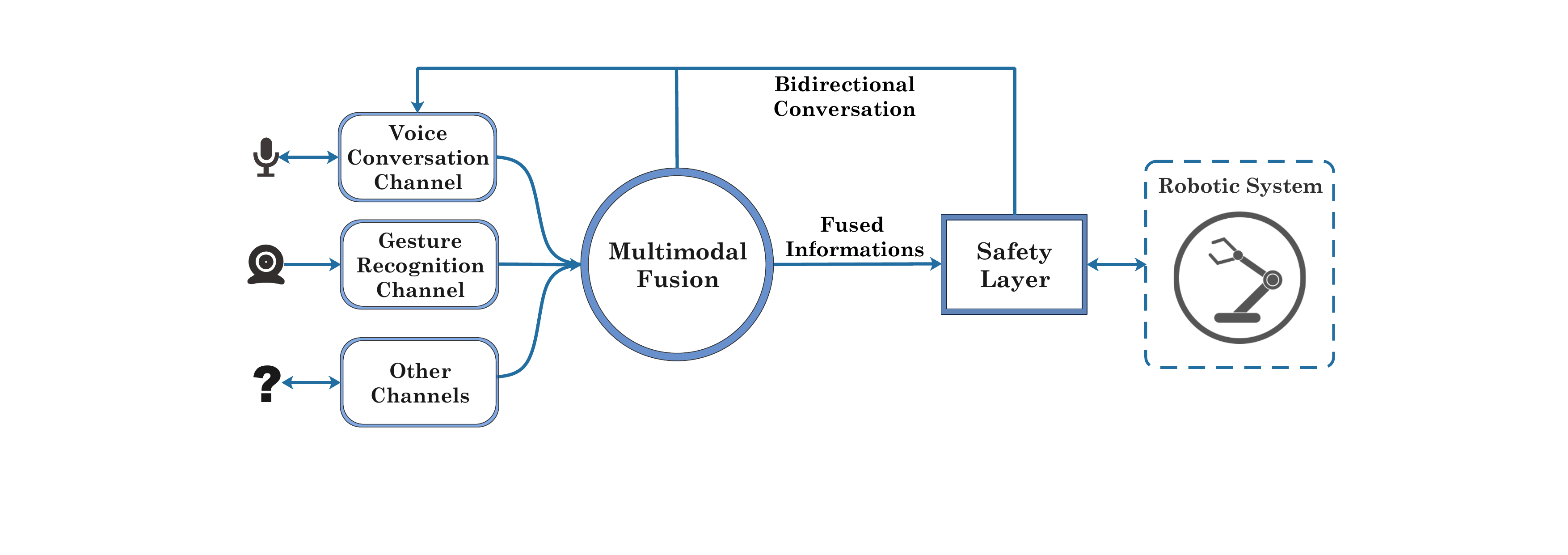}
        \caption{Proposed Architecture}
        \label{fig:multimodal architecture}
    \end{figure}

    The proposed architecture, depicted in Fig. \ref{fig:multimodal architecture}, integrates multiple communication channels, both unidirectional (e.g., gesture recognition) and bidirectional (e.g., voice conversation), contributing inputs to a multimodal fusion core.
    This core receives and synchronizes information from diverse channels, generating hybrid data to reconstruct the received multimodal communication. Additionally, a crucial safety layer manages robot inputs, simulates potential issues through a simulation algorithm, communicates errors, obstacles, or blockages, and proactively initiates conversations to address potential hazards.
    Our Human-Robot Collaboration (HRC) approach focuses on facilitating natural multimodal conversations, resembling human-to-human dialogue fluidity, and preventing the activation of constraints imposed by ISO safety standards through communication. In our architectural framework, communication channels can be interchangeably used by both humans and robots to initiate conversations, exchange information, coordinate tasks, and make requests, fostering a dynamic interaction model.
    To enable bidirectional communication, we implemented a core multimodal fusion system that seamlessly integrates various communication channels, such as voice and gestures, creating an efficient and versatile communication environment. Safety standards are seamlessly integrated within this communicative framework to proactively address potential safety issues. A predictive simulator anticipates potential slowdowns or blockages in the robot's trajectory, engaging in proactive dialogue with the operator to preemptively address safety concerns, prevent disruptions, and ensure smoother workflows while minimizing risks.
    Our architecture uniquely combines effective communication and safety standards as integral components, distinguishing it from existing approaches \cite{Ferrari2022ICRA}. Unlike traditional methodologies that often treat safety as a lower-level layer, we elevate it to an equal standing with communication, ensuring compliance with ISO/TS 15066 safety standards. This approach establishes safety as a core component of our communication strategies with operators.

    \subsection{Multimodal Fusion}

        Our multimodal fusion approach is grounded in the architecture outlined in \cite{Ferrari2023HFR}. It utilizes a combination of diverse communication channels seamlessly merged through a multimodal fusion algorithm. The primary objective of this integration is to form a unified and comprehensive representation of the communicated message, leveraging inputs from various communication channels.
        The process of multimodal fusion involves two key steps. Initially, information originating from these communication channels is collected and managed by a time manager inspired by the recognition lines discussed in \cite{cacace2017roman}. This crucial component is responsible for synchronizing and consolidating data from different sources into a singular tensor.
        Following this, a neural network classifier processes the tensor and generates the fused multimodal command, which is essentially a coherent and holistic representation of the received communication.
        The decision to employ a neural network as the classifier is driven by its intrinsic ability to learn and adapt to the complex, non-linear nature of multimodal data. Compared to other types of classifiers, neural networks often demonstrate greater flexibility in modeling intricate relationships and capturing complex patterns in data. This adaptability is particularly advantageous when handling signals from diverse communication channels. Furthermore, their adaptability allows seamless integration of new inputs and channels through the incorporation of additional training data. This flexibility ensures the system's responsiveness to evolving communication scenarios, making it a versatile and effective choice for real-world applications.
        As outlined in Algorithm \ref{algo:Multimodal-Fusion}, upon the arrival of new information from either the voice or gesture channel, a temporal window $\mathcal{W}$ is initiated. Within this window, any additional information received by each communication channel, denoted as $(X_1, X_2, \ldots, X_n)$, is concatenated into a tensor $\mathcal{T} = [X_1, X_2, \ldots, X_n]$ of predefined length, applying the zero-padding technique to fill the available slots for each communication channel and ensuring conformity with the input dimension of the neural network.
        Subsequently, the synchronized tensor $\mathcal{T}$ is given as input to a pre-trained neural classifier $\mathcal{N}$, represented by a function $\mathcal{M} = F(\mathcal{T})$, which generates a fused multimodal command $\mathcal{M}$. 

        This neural network is structured as a sequential model consisting of three linear layers interspersed with Rectified Linear Unit (ReLU) \cite{agarap2018deep} activation function $\sigma(x)=\max(0,x)$ that introduces non-linearity into the model, allowing it to learn complex patterns in the data. The network is parameterized by a set of weights $W$ and biases $b$, which are fine-tuned during the training process. The classification function can be expressed as:
        \begin{equation}
            \mathcal{M} = \sigma(W \cdot \mathcal{T} + b)
            \label{eq: classification function}
        \end{equation}

        \noindent
        The input and output sizes are dependent on the nature of the communication channels and the experimental setup. In our case, the input size is set to 4, reflecting the constraint of a narrow temporal window where it is unlikely to receive more than one vocal and gestural command and the output size is configured to 15 classes, each corresponding to a specific meaning in the context of the experiment, such as \textit{"place object there"}, \textit{"re-plan trajectory}, etc. Additionally, each hidden layer has a hidden size of 64 neurons, contributing to the network's capacity to capture complex relationships within the data.

        \begin{algorithm}[htbp]
            \small
            \begin{algorithmic}[1]
                \Require Vectors $G$ (\textit{Gesture}) and $V$ (\textit{Voice}) information
                \Ensure Multimodal Command $\mathcal{M}$
                \State $\mathcal{T} \gets$ empty tensor
                \State Recognition Time: $R_T \gets 2s$
                \If {new value of $G$ or $V$ is received}
                    \State Open a \textit{Temporal Window} $\mathcal{W}$
                    \State $\mathcal{T} \gets$ Received Value ($G$ or $V$)
                    \While {$\mathcal{W} < R_T$}
                        \If {new gesture/voice $G$ or $V$ is received}
                            \State Fill tensor $\mathcal{T}$ with new $G$ or $V$
                        \EndIf
                    \EndWhile
                    \State Pass $\mathcal{T}$ in Classifier $\mathcal{N}$ $\rightarrow$ Multimodal Command $\mathcal{M}$
                    \State Send $\mathcal{M}$ to the Safety Layer
                \EndIf
            \end{algorithmic}
            \caption{Multimodal Fusion Algorithm}
            \label{algo:Multimodal-Fusion}
        \end{algorithm}

        


        We constructed a custom dataset collected experimentally by classifying various combinations of inputs into classes based on their meanings in the context of the experiment. Each data point in the dataset is represented as $(\mathcal{T}_i, Y_i)$, where $\mathcal{T}_i$ corresponds to the synchronized tensor from multiple communication channels, and $Y_i$ is the corresponding ground truth characterizing the desired fused representation.
        The training process was conducted using the Stochastic Gradient Descent (SGD) optimization algorithm with a batch size of 64 and a learning rate of 4e-3 to minimize a loss function $L$ that quantifies the difference between the predicted $\mathcal{M}$ and the ground truth $Y$:
        
        \begin{equation}
            L = \sum_i \mathcal{L}(\mathcal{M}_i, Y_i),
            \label{eq:SGD equation}
        \end{equation}
        where $\mathcal{L}$ is a suitable loss function, such as cross-entropy. Additionally, we implemented the early-stopping technique with a patience of 100 epochs to monitor the trend of the loss function and stop the training prematurely to prevent overfitting.


    \subsection{Safety}

        Safety is maintained through a dedicated layer, outlined in Sec. \ref{sec:safety_layer}, which is responsible for both protecting the human operator and avoid undesired slowdowns. The ISO/TS 15066 sets a maximum limit on the relative human-robot speed, as indicated in equation \eqref{eq:vellimit}. However, adhering to this constraint might result in a too conservative robot behaviour. Thanks to the bidirectional communication channel, the robot can communicate its intentions to the human operator, preventing undesired speed reduction and enhancing the overall collaboration performance.

    \section{Safety Layer}\label{sec:safety_layer}
The safety layer has two main goals. Firstly, it ensures that the robot behavior complies with safety standards. Secondly, it anticipates and communicates future speed reductions to improve performance. 
Thanks to the proposed approach, the robot is capable of warning the human operator about its intention, emulating real human-human communication. Indeed, when two human operators collaborate, it is common for one of the two to ask the other to move since it has to reach the same area of the workspace.  This translates into an improvement of the mutual communication and comprehension of the two different agents, without compromising safety.
To achieve this, the safety layer computes for each task a collision-free trajectory $\traj$, which is decomposed through a path-velocity decomposition as detailed in \eqref{eq:q_s}-\eqref{eq:q_s_dot}. Subsequently, the trajectory is forwarded to two different components, each of which is responsible for one of the goals: online safety and predictive simulator.

\subsection{Online Safety}
The online safety of the human operator is guaranteed by a velocity scaling algorithm which was initially proposed in \cite{pupa2021safety}. In particular, starting from the trajectory, the safety layer solves 
 online the following optimization problem:
    \begin{equation}
        \label{eq:problem}
        \begin{array}{ll@{}ll}
            \displaystyle \min_{\alpha} & -\alpha,                                                                                                                      \\[0.2cm]
            \text{s.t.}                                                                                                                                               \\[0.2cm]
            & \displaystyle J_{r_i}({\q})\dqs\dot{s}\alpha \leq v_{max_i}        \quad \forall i \in \{1,\dots,n\}, \\[0.2cm]
            & \displaystyle \dq_{min} \leq \dqs\dot{s}\alpha \leq \dq_{max},                                                                \\[0.2cm]
            & \displaystyle \ddot{\q}_{min}\leq \dfrac{\dqs\dot{s}\alpha - \dq}{T_r} \leq \ddot{\q}_{max},                                   \\[0.2cm]
            & \displaystyle 0 \leq \alpha \leq 1.                                                                                           \\[0.2cm]
        \end{array}
    \end{equation}
    $\alpha \in \begin{bmatrix}0, 1 \end{bmatrix}$ is the optimization variable and represents the scaling factor.  $J_{r_i}(\q)\in \mathbb{R}^{1 \times n}$ is a \textit{modified jacobian} that takes into account only the scalar velocity of the $i$-th link towards the human operator, see \cite{pupa2021safety}. $v_{max_i}$ is the velocity limit imposed by the ISO/TS 15066 \cite{isots}. 
    $\dq_{min} \in \mathbb{R}^{n}$ and $\dq_{max} \in \mathbb{R}^{n}$ are the joint velocity lower bounds and the joint velocity upper bounds, respectively. While $\ddot{\q}_{min} \in \mathbb{R}^{n}$
    and $\ddot{\q}_{max} \in \mathbb{R}^{n}$ are the acceleration limits. $\dq \in \mathbb{R}^{n}$ is the actual robot velocity and $T_r$ is the robot execution time.

\subsection{Predictive Simulator}
    The predictive simulator has the goal of predicting and avoiding undesired modulation of the speed. Indeed, according to  \eqref{eq:problem}, when the human operator and the robot are close and the robot is going towards the human operator, the robot speed is scaled over the path to ensure compliance with the safety standards.
    However, these situations may cause a stop of the robot for a huge time, i.e. $\alpha=0$ until the human operator moves away.
    Thus, it would be more beneficial to communicate that the robot will decrease its speed so that the human operator can decide to move away and avoid useless stuck.

    The predictive simulator strategy is implemented according to Algorithm \ref{alg:srp}.
       \begin{algorithm}
       \small
            \begin{algorithmic}[1]
                \Require $\trajs, \dtrajs$
                \State $triggered \leftarrow false$ \label{algl:tfalse}
                \State $\q_{end}\leftarrow getEnd(\trajs)$ \label{algl:qend}
                \While{not $triggered$}
                    \State $\q_{virt}\leftarrow getRealState(\trajs)$
                    \State $End_{traj}\leftarrow false$
                    \State $T_{virt}\leftarrow 0$
                    \State $T_{rem} \leftarrow getDuration(\trajs, \q_{virt})$
                    \While{not $End_{traj}$}
                        \State $\dot{s}\leftarrow getSpeed(\dtrajs, \q_{virt})$
                        \State $H_{info}\leftarrow getHumanData()$
                        \State $\alpha_{virt}\leftarrow solveOpt(\dot{s}, \q_{virt}, H_{info})$
                        \State $\q_{virt} \leftarrow integrate(\dot{s}, \q_{virt}, \alpha_{virt})$ \label{algl:integration}
                        \State $T_{virt}\leftarrow T_{virt} + T_r$
                        \If{$\q_{virt} = \q_{end}$}
                            \State $End_{traj} \leftarrow true$ \label{algl:trajend}
                        \EndIf
                        \If{$checkTime(T_{virt},T_{rem})$} \label{algl:check}
                            \State $sendSignal()$\label{algl:signal}
                            \State $triggered\leftarrow true$\label{algl:ttrue}
                            \State \textbf{break}
                        \EndIf
                    \EndWhile
                \EndWhile
            \end{algorithmic}
            \caption{Predictive Simulator}
            \label{alg:srp}
        \end{algorithm}  
        The algorithm requires as input the parametrized trajectory and its derivative, namely $\trajs$ and $\dtrajs$. It immediately sets to $false$ the variable $triggered$, which indicates if the warning message has been sent, and finds what is the last robot configuration along the path, i.e. $\q_{end}$ (Lines \ref{algl:tfalse}-\ref{algl:qend}). Subsequently, it starts a while loop in which it initially reads the real robot position along the planned path and initializes the state of the virtual robot $\q_{virt}$. Then, the algorithm initializes the variable $End_{traj}$, which is used to check if the virtual robot has concluded the planned path, the virtual time $T_{virt}$, and it computes the remaining trajectory time $T_{rem}$, i.e. the ideal time that the robot needs to conclude the trajectory. Then, the algorithm starts an inner while loop. In this loop, the predictive simulator continuously updates the information regarding the human operator, i.e. position, velocity and human-robot distance. This information can be computed by exploiting standards techniques already available in the literature, e.g. distance between capsules \cite{ferraguti2020safety} and human motion prediction \cite{liu2017human}. Subsequently, the algorithm computes the optimal scaling factor of the virtual robot $\alpha_{virt}$ by solving the problem in \eqref{eq:problem} and integrates the dynamics of the virtual robot (Line \ref{algl:integration}). Such integration can be achieved by exploiting standard methods, e.g. forward Euler \cite{biswas2013discussion}. Then, the algorithm increments $T_{virt}$ and checks if the virtual robot has reached the end of the path. If this is the case, the predictive simulator exits from the inner while and restarts the check from the actual robot configuration (Line \ref{algl:trajend}).
        If the path is not concluded, the algorithm checks if $T_{virt}$ significantly exceeds the remaining time (Line \ref{algl:check}). If this is the case, it means that the robot will have to reduce the speed too much, with a consequent reduction in the performance. Therefore, the algorithm communicates the warning message and stops (Lines \ref{algl:signal}-\ref{algl:ttrue}). 
        The algorithm is activated every time the safety layer computes a new trajectory, this is because if the human operator decides to not move it is not necessary to communicate again possible speed reduction. 
        
        It is worth underline that, to achieve its functionality,the predictive simulator needs to operate at a frequency much higher than the robot controller. This is doable because the robot is modeled as a kinematic system, as detailed in Sec. \ref{sec:problem_statement}.

    \section{Experimental Validation}\label{sec:exp}

\begin{figure}
        \vspace{5pt}
        \centering
        \includegraphics[width=0.8\linewidth]{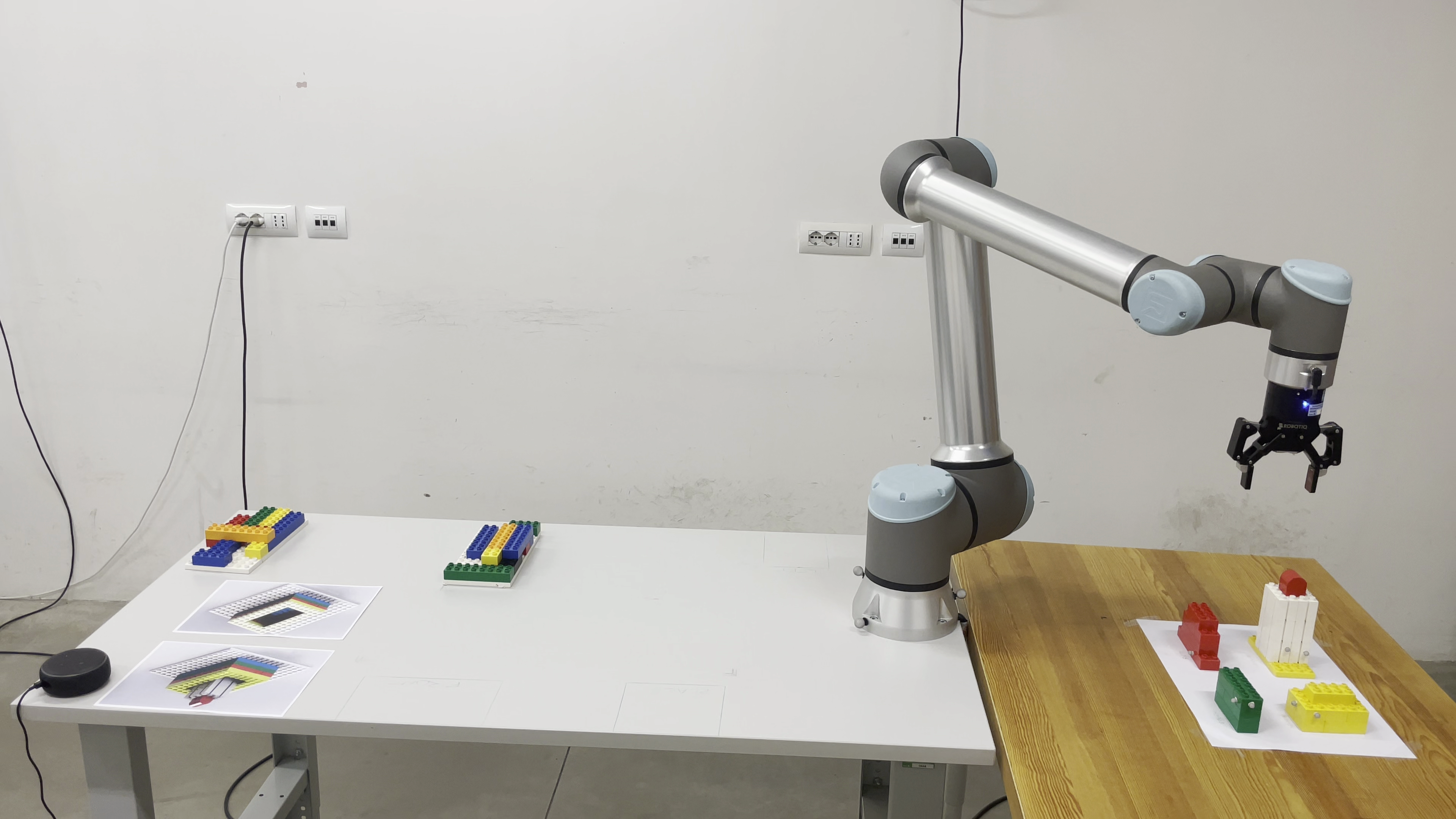}
        \caption{Setup of the Experiment}
        \label{fig:setup}
        \vspace{-3pt}
    \end{figure}

    The experimental validation\footnote{In the accompanying video, a sneak peek of some key parts of the experiment is showcased.} 
    of our bidirectional multimodal communication control architecture involved a comparative study between the proposed architecture and a state-of-the-art bidirectional communication system by Ferrari et al. \cite{Ferrari2022ICRA}, which simply adheres to ISO safety standards through a low-level layer.
    The study involved 12 participants, aged between 20 and 30 years, with an equal gender distribution and varying levels of experience with robotic systems, ranging from first-time encounters to several years of experience with collaborative robots, in order to ensure a representative sample. To mitigate potential learning biases, each participant performed both versions of the experiment, and the execution order was randomly determined for each individual, minimizing the impact of experiment repetition on the results.
    In this experimental scenario (Fig. \ref{fig:setup}), participants engaged in a collaborative assembly task where they assembled a set of LEGO components. The robot, a UR10e collaborative manipulator, played a supportive role by providing the necessary components to the operator. The experiments encompassed a variety of interactions, including requests for assistance from the robot, error notifications, and anticipation of potential issues. Central to our experiment was the communication between the robot and the operator, facilitated through both vocal conversation and a 3D gesture recognition channel. The robot can initiate conversations with the operator to communicate events and work together to find or propose solutions to emerging challenges. Conversely, the operator can send input to the robot through both the vocal conversation and gesture recognition channels. The multimodal fusion algorithm fused information from the multiple communication channels, enabling the creation of commands like \textit{``place the object in that area''} with the operator indicating the desired area through a \textit{``Point-At''} gesture, extrapolating the target zone by interpolating the coordinates of the elbow, wrist, and finger using a skeletonization algorithm. The primary objective of these experiments was to evaluate the effectiveness of multimodal conversation compared to common bidirectional communication while ensuring compliance with safety regulations. Additionally, we assessed its impact on collaboration efficiency and safety, including a predictive simulator to predict and preemptively address potentially hazardous situations through communication with the operator. Our goal was to highlight the advantages and differences brought about by our innovative approach.

    The following sections present a detailed discussion of the architecture implementation and the experimental results, providing insights into the benefits and enhancements achieved by our multimodal conversation architecture.

    \subsection{Implementation Details}\label{subsec:implementation}

        The architecture was developed using the \textit{ROS} framework \cite{ros}, with components organized into independent nodes to ensure modularity. For the voice communication channel, we created a custom \textit{Amazon Alexa skill}, designed using \textit{Alexa Conversations} \cite{acharya2021alexa}, a Deep Learning-based approach that employs API calls to manage multi-turn dialogues between Alexa and the user, resulting in more natural and human-like interactions. The skill's back-end was locally developed in Python (non-Alexa-hosted), enabling seamless integration with ROS by utilizing \textit{Microsoft Azure's HTTP Trigger Functions}. We employed \textit{ngrok} to expose the back-end and establish a connection to the front-end via HTTPS tunneling. To enable bidirectional communication and allow the robot to initiate conversations, we integrated \textit{Node-RED} \cite{nodered}, a web service for logical path programming. \textit{Node-RED} provides direct interaction with Alexa APIs, empowering the robot to report errors, manage events, and start conversations by invoking specific dialogue APIs, thereby facilitating the exchange of information and task execution.
        The gesture recognition channel was established using the \textit{Holistic} landmarks detection solution API from \textit{MediaPipe} \cite{lugaresi2019mediapipe}, a framework developed by Google that offers a comprehensive suite of libraries and tools for the application of artificial intelligence (AI) and machine learning (ML) techniques. Holistic combines elements of pose, face, and hand landmarks to create a unified landmarking system for the human body, operating in real-time on a continuous stream of images. The landmarks extracted from each image are encoded into a tensor to represent 3D gestures, which are then fed into a neural network classifier. This classifier consists of an LSTM (Long Short-term Memory) layer, followed by several fully connected layers, designed to classify gestures based on the extracted landmarks. The neural network model was trained using a dataset of communicative gestures specifically curated for human-robot collaboration \cite{Tan2021ArXiv}.

    \subsection{Analysis of the Results}\label{subsec:results}

\begin{figure}
        \vspace{5pt}
            \centering
            \includegraphics[width=\linewidth]{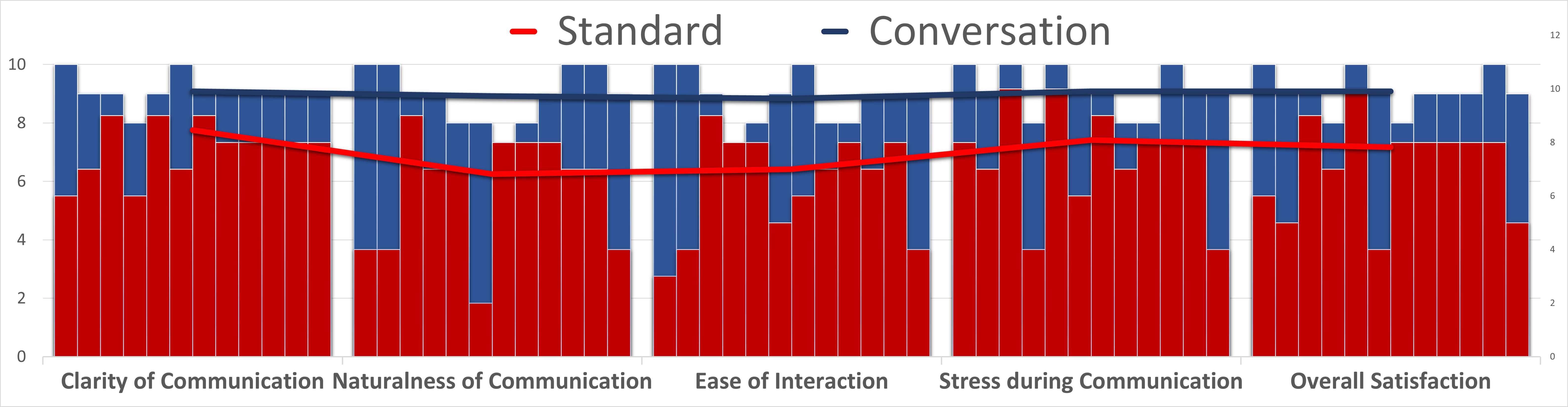}
            \caption{Questionnaire Results}
            \label{fig:questionnaire}
        \end{figure}

        To assess the effectiveness of the architecture, we measured the execution times and downtime of the robot during both versions of the experiment. Given that compliance with ISO safety standards is ensured in both experiments, we aim to use these metrics to valuate how well our architecture can predict and prevent slowdowns or safety stops through communication, specifically those arising from excessive proximity to the operator.
        Additionally, for each experiment, we administered a questionnaire to the participants, consisting in five ratings on a scale from 0 to 10, covering \textit{Clarity of Communication}, \textit{Naturalness of Communication}, \textit{Ease of Interaction}, \textit{Stress during Communication}, and \textit{Overall Satisfaction}.
        The graph in Fig. \ref{fig:questionnaire} displays the questionnaire results, with values from the comparative experiment shown in red and those from the architecture proposed in this article shown in blue. The results indicate a significant difference: the proposed architecture has an average score of approximately 9/10, while Ferrari at al. \cite{Ferrari2022ICRA} approach averages around 7/10. This suggests a notable improvement in the user experience when using the proposed architecture, both in terms of clarity and ease of use, as well as in terms of reduced stress due to more natural communication.
        
        \begin{figure}
            \minipage{0.49\columnwidth}
            \centering
                \includegraphics[trim={0 0 0 0}, clip, width=\linewidth]{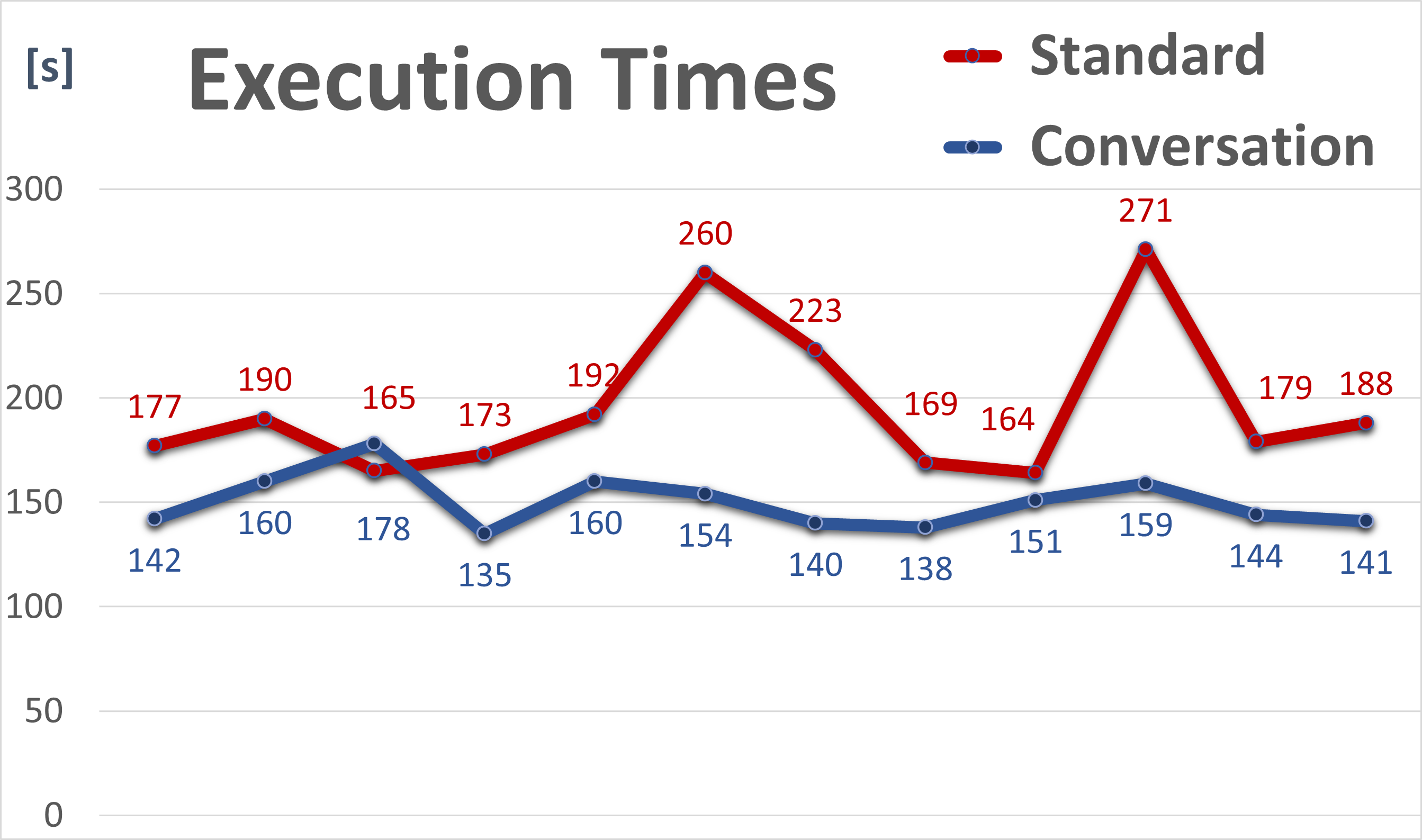}
            \endminipage\hfill
            \minipage{0.49\columnwidth}
            \centering
                \includegraphics[trim={0 0 0 0}, clip, width=\linewidth]{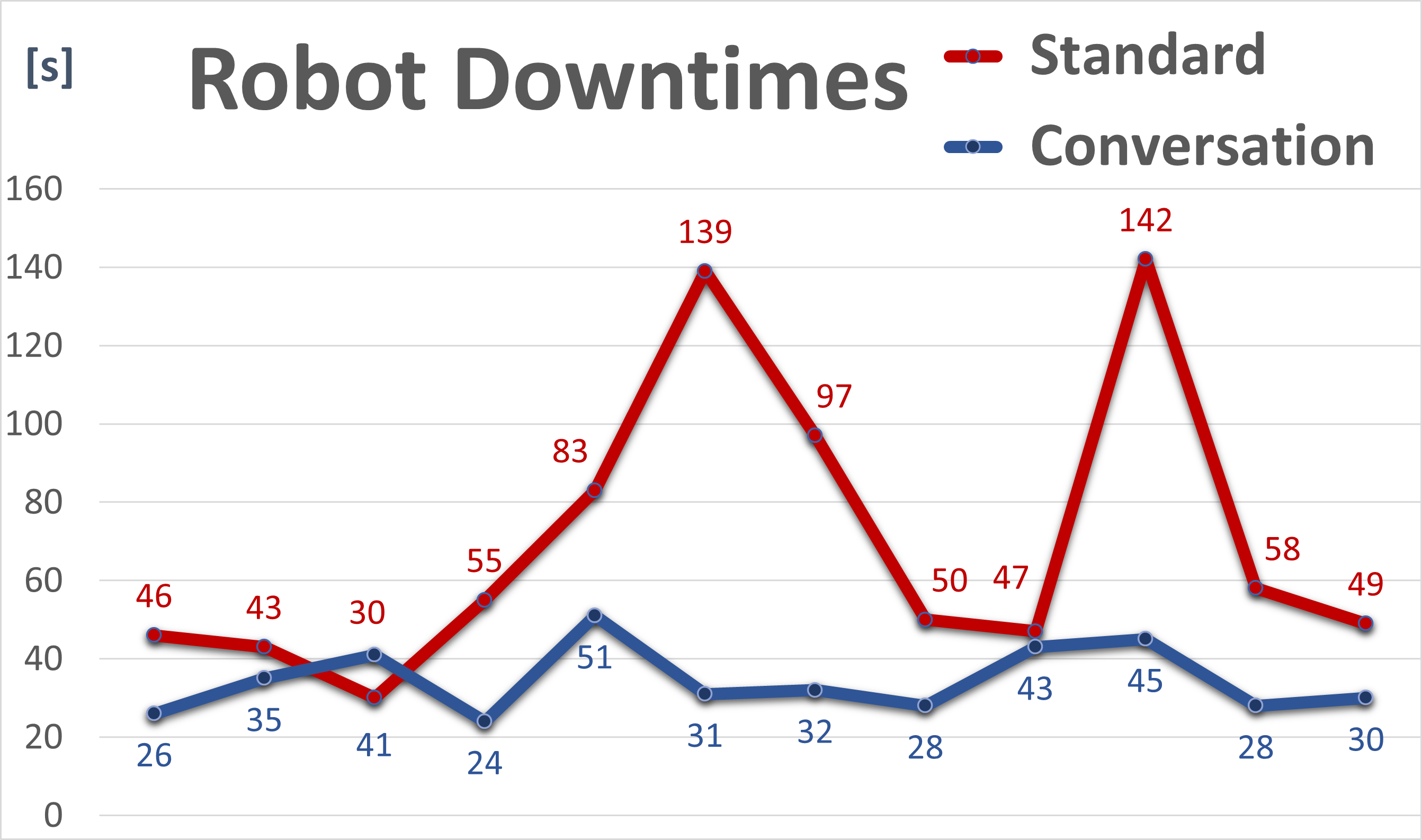}
            \endminipage\hfill
            \caption{Execution Times and Robot Downtime}
            \label{fig:results-time}
        \end{figure}

        The results of the experiment \textit{Execution Times} and \textit{Robot downtime} are depicted in Fig. \ref{fig:results-time}. It is evident that the utilization of the proposed architecture leads to a significant enhancement in job performance, with an average completion time of 150 seconds compared to the standard communication's average of 196 seconds. Furthermore, robot downtime also dramatically decreases with the use of the proposed architecture, reducing from an average of 70 seconds to 35 seconds. Thanks to the integrated predictive simulator and communication, potential blockages and slowdowns were anticipated and avoided, leading to faster execution of the required tasks and reduced robot downtime.
        \begin{table}
            \vspace{5pt}
            \small
            \caption{ANOVA Summary and Results - Execution Times}
            \label{tab:ANOVA_execution_times}
            \centering
            \resizebox{\linewidth}{!}{
            \begin{tabular}{c|c|c|c|c}
                \toprule
                \textbf{Groups} & \textbf{Count} & \textbf{Sum} & \textbf{Average} & \textbf{Variance} \\
                \midrule
                Standard Communication & 12 & 2351 & 195.92 & 1314.45 \\
                Multimodal Conversation & 12 & 1802 & 150.17 & 157.42 \\
                \bottomrule
                \toprule
                \textbf{F} & \multicolumn{2}{c|}{\textbf{P-value}} & \multicolumn{2}{c}{\textbf{F crit.}} \\
                \midrule
                17.065 & \multicolumn{2}{c|}{0.00044} & \multicolumn{2}{c}{4.30095} \\
				\bottomrule
            \end{tabular}}
        \end{table}
        \begin{table}[t]
            \vspace{5pt}
            \caption{ANOVA Summary and Results - Robot Downtime}
            \label{tab:ANOVA_downtime}
            \centering
            \resizebox{\linewidth}{!}{
            \begin{tabular}{c|c|c|c|c}
                \toprule
                \textbf{Groups} & \textbf{Count} & \textbf{Sum} & \textbf{Average} & \textbf{Variance} \\
                \midrule
                Standard Communication & 12 & 839 & 69.92 & 1407.90 \\
                Multimodal Conversation & 12 & 414 & 34.5 & 73 \\
                \bottomrule
                \toprule
                \textbf{F} & \multicolumn{2}{c|}{\textbf{P-value}} & \multicolumn{2}{c}{\textbf{F crit.}} \\
                \midrule
                10.164 & \multicolumn{2}{c|}{0.00425} & \multicolumn{2}{c}{4.30095} \\
				\bottomrule
            \end{tabular}}
        \end{table}
        \begin{table}[H]
            \vspace{5pt}
            \caption{ANOVA Summary and Results - Questionnaire Results}
            \label{tab:ANOVA_questionnaire}
            \centering
            \resizebox{\linewidth}{!}{
            \begin{tabular}{c|c|c|c|c}
                \toprule
                \textbf{Groups} & \textbf{Count} & \textbf{Sum} & \textbf{Average} & \textbf{Variance} \\
                \midrule
                Standard Communication & 5 & 35 & 7 & 0.4167 \\
                Multimodal Conversation & 5 & 45 & 9 & 0.0139 \\
                \bottomrule
                \toprule
                \textbf{F} & \multicolumn{2}{c|}{\textbf{P-value}} & \multicolumn{2}{c}{\textbf{F crit.}} \\
                \midrule
                46.452 & \multicolumn{2}{c|}{0.00014} & \multicolumn{2}{c}{5.31765} \\
				\bottomrule
            \end{tabular}}
        \end{table}
        To evaluate the statistical significance of the experiment and validate our findings, we conducted a single-factor ANOVA test for execution times, robot downtime and questionnaire results. Tables \ref{tab:ANOVA_execution_times}, \ref{tab:ANOVA_downtime} and \ref{tab:ANOVA_questionnaire} summarize the collected data, including mean values, sums, and variances, categorized by the two experiments. The results demonstrate that in all the ANOVA tests, the calculated F-value is significantly higher than the F-critical value, and the p-value is well below the significance level (alpha = 0.05), confirming a statistically significant difference between the two experiments.

    \section{Conclusions and Future Work}\label{sec:conclusions}


    In this paper, we have presented a novel bidirectional multimodal communication architecture designed to enhance human-robot collaboration in shared workspaces while prioritizing safety. Our architecture enables robots and human operators to engage in natural and effective conversations, mirroring the fluidity of human-to-human dialogue while adhering to ISO safety standards. 
    The experimental validation demonstrates the effectiveness of the architecture, achieving great collaboration efficiency and a more user-friendly and natural interaction experience.
    
    Future works will aim at expand the communication capabilities of our architecture by introducing new communication channels and modalities, allowing for even richer and more versatile interactions between humans and robots.
    Subsequently, we plan to incorporate vision AI algorithms to enhance error handling and object detection. This addition will empower robots to better perceive and react to their environment.
    Furthermore, it is possible to implement a real-time user monitoring system capable of assessing the operator's status, stress levels, and concentration. This monitoring will enable the robot to adapt its behavior to better assist the operator and proactively address any emerging issues. 
    

    \bibliographystyle{IEEEtran}
    \bibliography{Bibliography.bib}

\end{document}